\documentclass{article}

\usepackage{arxiv}

\usepackage[utf8]{inputenc} % allow utf-8 input
\usepackage[T1]{fontenc}    % use 8-bit T1 fonts
\usepackage{hyperref}       % hyperlinks
\usepackage{url}            % simple URL typesetting
\usepackage{booktabs}       % professional-quality tables
\usepackage{amsfonts}       % blackboard math symbols
\usepackage{nicefrac}       % compact symbols for 1/2, etc.
\usepackage{microtype}      % microtypography
\usepackage{lipsum}
\usepackage{authblk}
\usepackage{graphicx}
\usepackage{amsmath}
\usepackage{comment}
\usepackage{xcolor}
\usepackage{adjustbox}
\usepackage{booktabs}
\usepackage{subcaption}
\usepackage{float}
\usepackage{algorithmic}
\usepackage{caption}

\title{Neural Network-based Vehicular Channel Estimation Performance: Effect of Noise in the Training Set}

\author[1,2,3]{\textbf{S.~A.~Ngorima}}
\author[1]{\textbf{A.~S.~J.~Helberg}}
\author[1,2,3]{\textbf{M.~H.~Davel}}

\affil[1]{Faculty of Engineering, North-West University, South Africa}
\affil[2]{Centre for Artificial Intelligence Research, South Africa}
\affil[3]{National Institute for Theoretical and Computational Sciences, South Africa}

\begin{document}

\maketitle

\begingroup
\renewcommand{\thefootnote}{}
\footnote{This work is a preprint of a published paper by the same name\cite{10.1007/978-3-031-78255-8_12}. The authenticated version is available online at \url{https://doi.org/10.1007/978-3-031-78255-8_12}.}
\endgroup

\begin{abstract}
Vehicular communication systems face significant challenges due to high mobility and rapidly changing environments, which affect the channel over which the signals travel. To address these challenges, neural network (NN)-based channel estimation methods have been suggested. These methods are primarily trained on high signal-to-noise ratio (SNR) with the assumption that training a NN in less noisy conditions can result in good generalisation. This study examines the effectiveness of training NN-based channel estimators on mixed SNR datasets compared to training solely on high SNR datasets, as seen in several related works. Estimators evaluated in this work include an architecture that uses convolutional layers and self-attention mechanisms; a method that employs temporal convolutional networks and data pilot-aided estimation; two methods that combine classical methods with multilayer perceptrons; and the current state-of-the-art model that combines Long-Short-Term Memory networks with data pilot-aided and temporal averaging methods as post processing. %Our results indicate that a training dataset with a high SNR is not always optimal, and the training SNR levels should be considered a hyperparameter to be adjusted. 
Our results indicate that using only high SNR data for training is not always optimal, and the SNR range in the training dataset should be treated as a hyperparameter that can be adjusted for better performance. This is illustrated by the better performance of some models in low SNR conditions when trained on the mixed SNR dataset, as opposed to when trained exclusively on high SNR data.
\end{abstract}

% keywords can be removed

\section{Introduction}
The integration of network infrastructure with artificial intelligence (AI) is becoming more prevalent due to the increasing demands of data-intensive applications and the need for more efficient and reliable communication systems. Incorporating AI into these systems promises to improve network management, resource allocation, and overall system performance.

This convergence is notably evident in vehicular communication, which is an important component of intelligent transportation systems. Vehicular communication systems face unique problems due to their high mobility, rapidly varying environments, and significant latency requirements. One of the challenges in this domain is obtaining accurate channel estimates, which is critical to ensuring reliable communication between vehicles and infrastructure. Channel estimation involves determining the characteristics of a communication channel and monitoring its changes to adjust to varying conditions~\cite{875230}.
%This is particularly evident in vehicular communication, where one of the most significant challenges is accurate channel estimation. 

The dynamic nature of vehicular environments, characterised by rapidly changing channel conditions, poses a challenge to traditional channel estimation methods. These methods include methods such as least squares (LS) and minimum mean square error (MMSE), which rely on mathematically estimating the channel based on the received version of known data sent as a preamble to the signal. Although effective in more stationary environments, these techniques often struggle to keep up with the rapid changes inherent in vehicular channels~\cite{gizzini2020deep}. This limitation highlights the need for more advanced adaptive approaches to channel estimation in these high-mobility scenarios. 

In recent years, researchers have increasingly focused on using data symbols directly to estimate the channel in a method referred to as data pilot-aided (DPA) channel estimation. DPA estimation addresses the challenges posed by dynamic environments, especially in vehicular communication systems, where pilot signals are evidently insufficient. Unlike traditional methods that rely heavily on a fixed number of pilot signals for channel estimation, DPA estimation uses demapped data symbols as additional pilots. This approach allows for more flexible and adaptive tracking of channel variations without the need to allocate more resources for pilot signals, thereby improving efficiency and accuracy in rapidly changing channels. However, the performance of the DPA method can be significantly affected by errors introduced during the demapping process, which may limit its overall effectiveness~\cite{pan2021channel}.
%However, the DPA method has a demapping error that significantly limits its performance~\cite{pan2021channel}. 
To improve efficiency, DPA estimation is typically used with an error compensation scheme to reduce channel distortion and prevent errors from propagating to the next symbols in a frame. Two common error compensation approaches for channel estimation are spectral temporal averaging (STA)~\cite{fernandez2011performance} which utilises the correlations of data symbols in time and frequency domains to estimate the channel and time-domain reliable test frequency domain interpolation (TRFI)~\cite{kim2014time}, as described in more detail in Section \ref{well-known}. 
However, these methods still struggle in highly dynamic scenarios~\cite{gizzini2021temporal}.

Deep learning (DL) has recently gained attention for its ability to improve the accuracy of channel estimates. A method utilising autoencoders and training on a high signal-to-noise ratio (SNR) dataset (SNR=40 dB) was proposed to correct DPA estimate errors in the frequency domain~\cite{han2019deep}. However, it did not account for temporal changes, limiting its performance. Furthermore, incorporating feedforward networks into classical methods such as STA and TRFI methods, as proposed in~\cite{gizzini2020deep}, has shown promise in improving estimation accuracy. These methods were also trained on high SNR datasets (SNR=30 dB).
 
These methods offer valuable baselines.
%Nevertheless, these methods are not effective on all signal-to-noise ratio SNRs. 
Recently, a long-short-term memory (LSTM)-based architecture has been proposed to capture the temporal and frequency characteristics of vehicular channels~\cite{gizzini2021temporal}. This method was trained on a dataset generated at the SNR level of 40 dB and demonstrated significant improvements over previous estimators. However, this method is computationally expensive. 

This study introduces the use of a mixed SNR dataset, where the training dataset includes data generated at different SNR levels. We hypothesise that training on a mixed SNR dataset helps an neural network (NN) generalise across a wider range of real-world conditions by exposing it to both noise-dominated and signal-dominated scenarios. This approach contrasts with the methodology adopted in several studies~\cite{gizzini2020deep,gizzini2020joint,gizzini2021temporal,pan2021channel}, which propose that training the NN-based estimator using a dataset generated only at a high SNR level enhances its ability to generalise in any environment: at high SNR the impact of the channel is greater than the impact of noise, therefore it is expected that the NN develops better channel knowledge. 

We evaluate the use of the mixed SNR dataset using a CNN-Transformer-based  estimator~\cite{Ngorima2024a} which is a hybrid model that combines a one-dimensional convolutional neural network (CNN) with a transformer architecture, a TCN-DPA estimator~\cite{Ngorima2024} that uses a temporal convolutional network (TCN) to correct the DPA propagation error, two estimators based on concatenating a multilayer perceptron (MLP) with a classical method, in this case STA-MLP~\cite{gizzini2020deep} and TRFI-MLP~\cite{gizzini2020joint}, and the LSTM-DPA-TA method~\cite{gizzini2021temporal} which combines Long Short-Term Memory (LSTM) networks with Data Pilot-Aided (DPA) estimation and Temporal Averaging (TA).

\section{Background}

We describe the system model, well-known channel estimation schemes in vehicular communications, and some of the NN-based channel estimation techniques currently used. 
We also provide a brief description of the IEEE 802.11p physical layer structure that is used in all simulations, along with a description of its specifications. 

\subsection{System Model}
\label{system_model}

\begin{figure}[ht]
    \centering
    \includegraphics[width=0.5\linewidth]{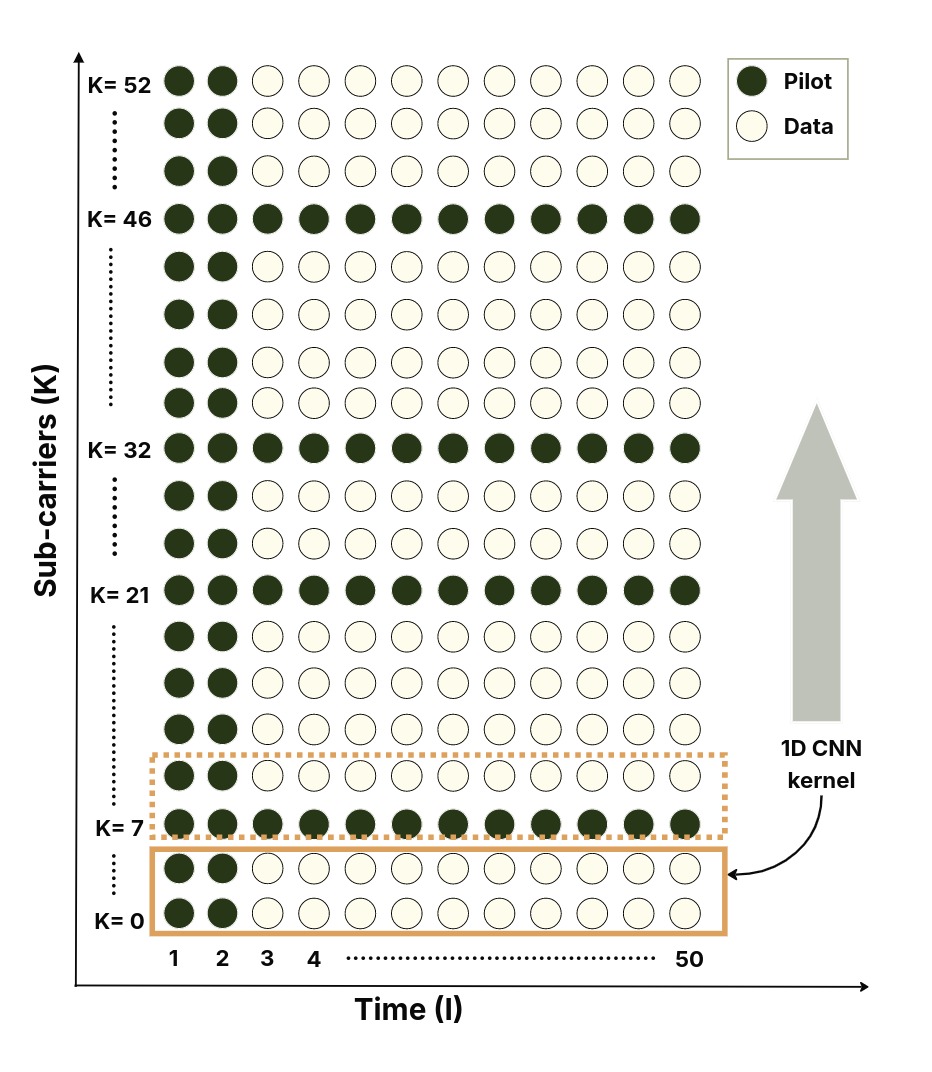}
    \caption{The IEEE 802.11p frame structure, illustrating the allocation of subcarriers for pilot and data transmission.}
    \label{fig:1DCNN}
\end{figure}

In this paper, we adopt the IEEE 802.11p physical layer structure, which is specifically designed for vehicular communication systems. This section follows a detailed discussion of the IEEE 802.11p specifications as outlined in Ngorima et al.~\cite{Ngorima2024}. %We also provide a detailed discussion of some of the channel estimation methods later used for this work's evaluation stage.
The IEEE 802.11p standard uses Orthogonal Frequency Division Multiplexing (OFDM) to transfer data by dividing the available bandwidth into several subcarrier frequencies for simultaneous transmission of multiple signals. 52 of the 64 available subcarriers are used for data transmission and pilot symbols. The remaining subcarriers are for guard bands and direct current (DC) offset. To illustrate this structure, Figure \ref{fig:1DCNN} shows the IEEE 802.11p OFDM frame, indicating pilot and data subcarriers.

In this work, we consider a frame structure that consists of two preamble symbols for signal detection and timing synchronisation, followed by a signal field that contains transmission parameters.
The received signal on subcarrier $[k]$ at time $i$ can be represented as
\begin{equation*} Y_{i}[k]=H_{i}[k]X_{i}[k]+N_{i}[k],\tag{1} \label{rec_signal}\end{equation*}
where $H_i[k]$, $X_i[k]$ and $N_{i}[k]$ denote the respective channel response, transmitted signal, and additive Gaussian noise (AWGN), respectively. Note that in vehicular channels $H_{i}$ varies in both the time and frequency domains.

\subsection{Well-known Vehicular Channel Estimation Schemes}
\label{well-known}
This section describes some of the main vehicular channel estimation schemes that have been widely used and studied in the literature. These methods were selected based on their effectiveness in dynamic environments and their foundational role in the development of more advanced NN-based estimators.
%The following are some of the vehicular channel estimation schemes. 

\subsubsection{DPA Estimation:}
The first step of the DPA process, as described by Pan et al.~\cite{pan2021channel}, is to calculate the initial estimate of the frame. This is done by applying the least squares (LS) estimation to the preamble as shown below:

% \subsubsection{DPA Estimation}

% The first step of the DPA process is to calculate the initial estimate of the frame. This is done by applying the least squares (LS) estimation to the preamble as shown below:

\begin{equation*} \hat { {h} }_{\text {0}}[k] = \frac { {y} ^{(p)}_{1}[k] + {y} ^{(p)}_{2}[k]}{2 {p} [k]}, \tag{2}\label{eqn:ls}\end{equation*}
where ${y} ^{(p)}_{1}[k]$ and ${y} ^{(p)}_{2}[k]$ are the known orthogonal preambles received on the $[k]$ subcarrier and $p[k]$ is the OFDM data symbol transmitted over the k-th subcarrier.
The next OFDM symbols in the frame are equalised using the initial estimate as the starting point as follows:
\begin{equation*} y_{i}^{\text{eq}}[k]= \frac{y_{i}[k]}{\hat{h}_{i-1}^{\text{DPA}}[k]},\hat{h}_{0}^{\text{DPA}}[k]=\hat{h}_{\text {0}}[k].\tag{3}\label{eqn:yeq}\end{equation*}
The equalised symbol is demapped and mapped to its closest constellation point as: %(see (\ref{eqn:conse})). 
\begin{equation*}d_{i}[k]=\mathfrak{R}\left(\frac{y_{i}[k]}{ \hat{h}_{i-1}^{\text{DPA}}[k]}\right), \hat{h}_{0}^{\text{DPA}}[k]=\hat{h}_{\text {LS}}[k],\tag{4}\label{eqn:conse}\end{equation*}
Finally, the DPA channel is updated: 
\begin{equation*}\tilde{h}_{\mathrm{D}\mathrm{P}\mathrm{A}_{i}}[k]=\frac{y_{i}[k]}{d_{i}[k]}\tag{5}\end{equation*}
\subsubsection{STA Estimation Scheme:}
STA uses both spectral (frequency) and temporal correlations within data symbols to estimate the channel~\cite{gizzini2020deep}. STA can also be viewed as a recursive filtering process that combines multiple channel estimates into a more accurate estimate.
The first step is to calculate a channel estimate in the frequency domain $\hat{h}{i}^{\text{FD}}[k]$ as follows: 
\begin{equation*} \hat{h}{i}^{\text{FD}}[k]=\sum{\lambda=-\beta}^{\beta}\omega_{\lambda}\hat{h}_{i}^{\text{DPA}}[k+\lambda],~\omega _{\lambda } = \frac {1}{2\beta +1},\tag{6}\end{equation*}
where $2\beta + 1$ represents the number of subcarriers that are considered.
Subsequently, a temporal average is computed to arrive at the final STA channel estimate, $\hat{h}{i}^{\text{STA}}[k]$:
\begin{equation*}\hat{h}{i}^{\text{STA}}[k]=\left(1-\frac{1}{\alpha}\right)\hat{h}{i-1}^{\text{STA}}[k]+\frac{1}{\alpha}\hat{h}_{i}^{\text{FD}}[k],\tag{7}\end{equation*}
where $\alpha$ is a smoothing parameter that controls the weight given to current and previous estimates.

\subsubsection{TRFI Estimation Scheme:}
TRFI leverages the high correlation within OFDM symbols to improve the accuracy of the channel estimation~\cite{gizzini2020joint}. The process starts by creating multiple channel estimates for each symbol using the DPA method. The received symbol is equalised using the estimates of the current and previous symbols ($\hat{h}_{i}^{\text{DPA}}[k]$) and ($\hat{h}_{i-1}^{\text{DPA}}[k]$) respectively.
\begin{align*}
{y}_{i-1}^{\text{eq}\prime}[k] &= \frac{y_{i-1}[k]}{\hat{h}_{i}^{\text{DPA}}[k]}, \\
{y}_{i-1}^{\text{eq}\prime\prime}[k] &= \frac{y_{i-1}[k]}{\hat{h}_{i-1}^{\text{DPA}}[k]}, \tag{8}\label{eqn:cdp_eq}
\end{align*}
Subsequently, the equalised symbols are remapped to their corresponding constellation points, ${d}_{i-1}^\prime[k]$ and ${d}_{i-1}^{\prime\prime}[k]$. 
TRFI then categorises subcarriers into reliable and unreliable sets based on a reliability test between the remapped symbols. Subcarriers with consistent remapped symbols, that is, ${d}_{i-1}^\prime[k]= {d}_{i-1}^{\prime\prime}[k]$, are considered reliable, while those with discrepancies are classified as unreliable. TRFI then interpolates channel estimates for unreliable subcarriers by using the reliable subcarriers as anchor points. This interpolation process effectively fills in the gaps in channel knowledge, resulting in a more accurate overall channel estimate. A detailed discussion on the interpolation procedure is given in~\cite{gizzini2020deep}.
\subsection{MLP-based Estimators} 
To enhance the performance of conventional STA and TRFI estimators, Gizzini et al.~\cite{gizzini2020deep} introduced MLP-based channel estimation methods. In these methods, MLP layers are added to the STA and TRFI processes, allowing the model to learn complex patterns in vehicular channels and thereby improve the accuracy of the estimation. The methods STA-MLP and TRFI-MLP initially conduct DPA estimation, proceed with STA or TRFI estimation and subsequently feed the results into the MLP layers. 
\subsection{TCN-DPA Estimator}
As proposed in~\cite{Ngorima2024}, the TCN-DPA estimator integrates a Temporal Convolutional Network (TCN) with DPA estimation to enhance channel estimation in dynamic environments. The TCN processes the received OFDM symbols in the frequency domain, treating subcarriers as time steps. The extracted features are then passed to the DPA process, where the previous TCN output, \(\hat{h}_{i-1}^{\text{TCN}}[k]\), is used to equalise the current received symbol. After equalisation, the symbol is demapped and remapped to the nearest constellation point to obtain the final channel estimate, \(\hat{h}_{i}^{\text{TCN}}[k]\). This estimate is updated iteratively using the DPA process. This approach leverages the ability of the TCN to model long-range dependencies across subcarriers, making it effective for complex channel conditions.

\subsection{LSTM-DPA-TA Estimator}
The LSTM-DPA-TA estimator~\cite{gizzini2021temporal} integrates an LSTM network with DPA and Temporal Averaging (TA) techniques to enhance channel estimation in vehicular environments. In this approach, the LSTM processes the received OFDM symbols to extract key features, which are then passed into the DPA module. The resulting channel estimates are further refined using the TA technique. In the LSTM-DPA-TA method, OFDM symbols are treated as time steps, with subcarriers considered as features within each time-step sequence. This design enables the model to effectively capture frequency dependencies and improve the accuracy of channel estimation under dynamic conditions.

\subsection{CNN-Transformer}
The CNN-Transformer is a recently proposed hybrid architecture for vehicular channel estimation in~\cite{Ngorima2024a}. This approach combines the strengths of CNNs and Transformer networks to analyse vehicular channel data comprehensively. The channel data can be represented in both the time and frequency domains. The frequency domain represents the subcarriers of the OFDM symbols, while the time domain represents the sequence of OFDM symbols transmitted over time. Each OFDM symbol is transmitted across multiple subcarriers. %, which are essentially distinct frequency bands. 
The frequency domain captures the spatial characteristics of the channel across OFDM subcarriers, while the time domain reflects the temporal dynamics as OFDM symbols are transmitted sequentially. 

In this CNN-Transformer architecture, subcarriers are treated as the `time steps'. 
The architecture leverages CNNs to extract local features from subcarriers and uses Transformer layers to capture global dependencies across the frequency domain. The CNN component applies 1D convolutions to capture local patterns, while the Transformer component uses self-attention mechanisms to analyse patterns between subcarriers globally.

\section{Methodology}
\label{proposed}

This section outlines the methodological framework used in this study.
%to assess the best training practices for different NN-based channel estimators. 
We start with a detailed description of the datasets used for training and evaluation, specifically focussing on two approaches: the mixed SNR dataset and the high SNR dataset. Following this, we discuss the data preprocessing steps applied to transform complex received symbols into a format suitable for input into NNs. 
\subsection{Data Preparation}
\label{dataprep}
We simulate vehicle-to-vehicle communication following the `Vehicle-To-Vehicle Expressway Same Direction with Wall' (VTV-SDWW) channel model~\cite{4526014}. Our simulation specifically modelled vehicles moving at 100 km/h, with a Doppler shift of 550 Hz. We used 16QAM (16-Quadrature Amplitude Modulation) modulation for data transmission in this urban environment scenario. This modulation scheme encodes the data by varying the amplitude and phase of the carrier signal.

\subsubsection{Mixed SNR Dataset}
The mixed SNR dataset represents a wide range of noise introduced on the VTV-SDWW channel. For this dataset, we simulate 18,000 time-specific frames with SNR values ranging from 0 to 40 dB in increments of 5 dB. Each SNR level includes 2,000 frames, with 50 OFDM symbols spread across 52 active subcarriers (48 data and 4 pilot subcarriers). During training, each model is exposed to the entire range of SNR levels.
The validation set is created by reserving 25\% of this training data to monitor the performance of the model during training. An independent test set consisting of 2,000 frames is generated separately from the training sets. This test set includes frames at SNR levels of 0, 5, 10, 15, 20, 25, 30, 35, and 40 dB, ensuring that the evaluation correctly reflects the performance across different SNR conditions. The same test set is used to test all the models regardless of how they are trained. 
\subsubsection{High SNR Dataset}
The high SNR dataset focusses on a fixed SNR level, representing less noisy or nearly ideal channel conditions, as opposed to the mixed SNR dataset. This dataset is used to train models with the assumption that training NNs in an environment with a clearly defined channel can lead to better generalisation even in noisy environments~\cite{gizzini2021temporal,gizzini2020deep,gizzini2020joint,pan2021channel}. 
In this work, the high SNR dataset consists of 18,000 time-specific frames, all generated at a 40 dB SNR level, with the same frame structure as the mixed SNR data. As before, 25\% of the training data is set aside for validation. 

\subsubsection{Data preprocessing}

\begin{figure}[ht] 
    \centering
    \includegraphics[width=\linewidth]{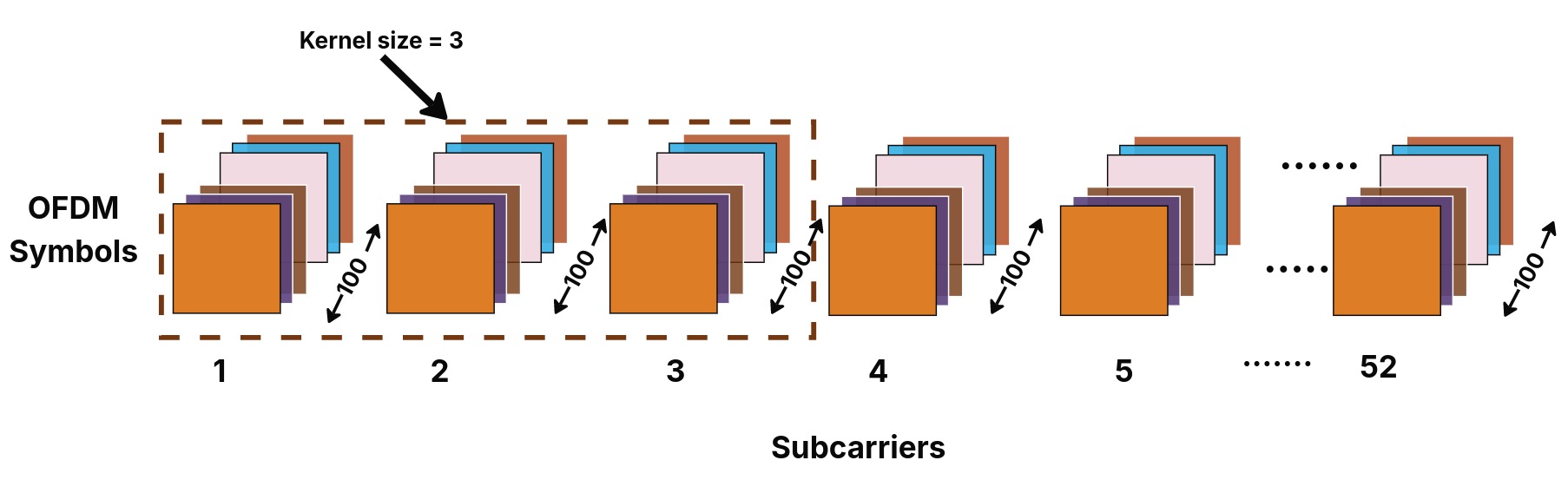}
    \caption{Transformed IEEE 802.11p frame structure used as input to deep learning models, composed of 100 interleaved complex symbols and 52 subcarriers.}
    \label{fig:frame}
\end{figure}

To preprocess the data for NN models, we transform the complex 16QAM symbols received across subcarriers and time slots into a real-valued format. Each received symbol consists of both real and imaginary components. The following steps are performed to prepare the data:

\begin{enumerate}
    \item \textbf{Separate Real and Imaginary Components:} Decompose each complex 16QAM symbol into its real and imaginary parts. Handle the data as two separate real-valued matrices instead of a single complex-valued matrix. For a given OFDM symbol, where \(y[k] = r[k] + j \cdot i[k] \) represents the received complex value at subcarrier \( k \), extract \( r[k] \) as the real part and \( i[k] \) as the imaginary part.

    \item \textbf{Interleave Components:} After separation, interleave the real and imaginary components. This means arranging the real and imaginary parts in an alternating sequence across the time slots. By interleaving, a unified structure that preserves the relationship between the real and imaginary parts within the input data is created.

    \item \textbf{Form the Input Matrix:} Structure the interleaved data into a matrix with dimensions \( 52 \times 100 \), where 52 represents the number of subcarriers and 100 represents the sequence of interleaved real and imaginary components across the time slots. This transformation enables the NN to process the complex-valued input as a series of real-valued inputs.

    \item \textbf{Prepare Input Data for NN models:} The resulting matrix is now a real-valued representation of the original complex data, ready for input into NN models. %This preprocessing step ensures that the network can effectively learn from the data without being hindered by the challenges of handling complex numbers directly.
\end{enumerate}
The transformed input matrix described above is visually represented in Figure~\ref{fig:frame}. Each block in the figure represents the interleaved real and imaginary components of the received 16QAM symbols for a particular subcarrier across the OFDM symbols. 
The kernel size of 3 shown in the figure indicates the width of the sliding window used by the CNN during the convolution process across the subcarriers.

\subsection{Hyperparameter Optimisation}
\label{sec:hyperparameters}
We reimplement most of the NN-based estimators discussed in the previous sections to ensure compatibility with our experimental setup and to be able to optimise each individually for the specific datasets used. Specifically, the CNN-Transformer, TCN-DPA, STA-MLP, and TRFI-MLP models were reimplemented from scratch. The LSTM-DPA-TA model is implemented using existing code available online. To verify the accuracy of our reimplementations, we compare the performance of our implementations against the benchmark results reported in the original studies, ensuring that they perform as expected.

The final architecture of each model is determined through a hyperparameter tuning process, to identify the optimal configurations for each model. We optimise the hyperparameters of all models in both datasets using Optuna, a Bayesian optimisation framework that efficiently searches the hyperparameter space using a tree-structured Parzen estimator (TPE)~\cite{akiba2019optuna}. This method systematically explores different hyperparameter configurations while conserving computational resources by early termination of underperforming trials. %The optimisation process covers a set of hyperparameters that address the important components of the model. %The steps in the optimisation procedure are as follows:

For each model, we conduct multiple trials with different random seeds to ensure that results are not biased by a particular initialisation. Specifically, we conduct 50 trials per model, each with 3 seeds, allowing us to assess the robustness of the selected hyperparameters. The search space for each model was designed to cover a wide range of potential configurations. The key hyperparameters include the learning rate, the number of layers, the layer size (kernel size,  number of attention heads, or layer width, depending on model), the learning rate scheduler parameters, and hyperparameters controlling regularisation through dropout. 
%The search space was determined on the basis of prior studies, domain knowledge, and preliminary experiments, ensuring that it was sufficiently broad to capture the optimal settings. 
During the optimisation process, we monitor the convergence of validation loss. Early stopping is employed to prevent overfitting, stopping training when the validation loss ceases to improve for a predefined number of epochs.

\begin{table}[hb!]
\caption{Optimised hyperparameters for CNN-Transformer, TCN-DPA, STA-MLP, TRFI-MLP, and LSTM-DPA-TA on the mixed SNR and high SNR datasets.}
\label{tab:all_hyperparameters}
\centering
\begin{tabular}{lccc}
\hline
\textbf{Hyperparameter} & \hspace{0.15em}\textbf{Search Space} & \hspace{0.15em}\textbf{Mixed SNR}\hspace{0.15em} & \hspace{0.25em}\textbf{40 dB}\hspace{1em} \\
\hline
\multicolumn{4}{l}{\textbf{CNN-Transformer}} \\
Learning rate & [1e-5 to 1e-2] & 0.001 & 0.001 \\
Transformer layers & [1 to 4] & 2 & 4 \\
Number of attention heads & [1 to 4] & 2 & 4 \\
Hidden dimension & [64 to 128] & 128 & 128 \\
Dropout rate & [0.001 to 0.3] & 0.1 & 0.25 \\
Number of epochs & [50 to 200] & 110 & 200 \\
Number of CNN layers & [1 to 5] & 2 & 4 \\
CNN kernel size & [2 to 5] & 3 & 3 \\
\hline
\multicolumn{4}{l}{\textbf{TCN-DPA}} \\
Learning rate & [1e-5 to 1e-2] & 0.0006 & 0.003 \\
Number of Layers & [1 to 5] & 4 & 4 \\
Kernel Size & [2 to 5] & 2 & 2 \\
Dropout & [$10^{-5}$ to 0.5] & 0.17 & 0.01 \\
StepLR Step Size & [10 to 50] & 21 & 17 \\
StepLR Gamma & [0.5 to 1] & 0.9 & 0.8 \\
Epochs & [0 to 200] & 156 & 100 \\
\hline
\multicolumn{4}{l}{\textbf{STA-MLP}} \\
Learning rate & [1e-5 to 1e-2] & 0.001 & 0.001 \\
Number of layers & [1 to 5] & 2 & 3 \\
Size of hidden layer 0 & [5 to 30] & 29 & 15 \\
Size of hidden layer 1 & [5 to 30] & 27 & 15 \\
Size of hidden layer 2 & [5 to 30] & N/A & 15 \\
Total training epochs & [50 to 500] & 133 & 300 \\
\hline
\multicolumn{4}{l}{\textbf{TRFI-MLP}} \\
Learning rate & [1e-5 to 1e-2] & 0.0004 & 0.001 \\
Number of layers & [1 to 5] & 3 & 3 \\
Size of hidden layer 0 & [5 to 30] & 23 & 15 \\
Size of hidden layer 1 & [5 to 30] & 29 & 15 \\
Size of hidden layer 2 & [5 to 30] & 21 & 15 \\
Total training epochs & [50 to 500] & 130 & 160 \\
\hline
\multicolumn{4}{l}{\textbf{LSTM-DPA-TA}} \\
Learning rate & [1e-5 to 1e-1] & 0.004 & 0.01 \\
LSTM size & [64 to 128] & 128 & 128 \\
StepLR step size & [1 to 50] & 35 & 10 \\
StepLR step gamma & [0.1 to 1] & 0.7 & 0.8 \\
Training epochs & [50 to 500] & 160 & 500 \\
\hline
\end{tabular}
\end{table}

Table~\ref{tab:all_hyperparameters} provides a summary of the best hyperparameters obtained for each model across the datasets. In addition to the parameters in the table, batch normalisation is applied to enhance model generalisation. The Adam optimiser and a batch size of 128 was used for TCN-DPA, LSTM-DPA-TA, STA-MLP and TRFI-MLP. For the CNN-Transformer model, we use the AdamW optimiser and manually set the batch size to 16.

\section{Results}
 This section evaluates the performance of channel estimators using BER as the primary metric. We compare the effectiveness of two training approaches: one using a mixed SNR dataset and the other using a high SNR dataset.  The evaluated architectures are CNN-Transformer, TCN-DPA, STA-MLP, TRFI-MLP, and LSTM-DPA-TA.

\subsection{BER}
BER is a crucial performance measure in digital communication. It compares transmitted and received bit sequences to assess system reliability. In channel estimation, a lower BER means more accurate signal reconstruction and fewer errors in decoded data. The following analysis details the BER performance of each model at different levels of SNR, providing a direct comparison between the two training approaches.
%\subsection{BER}
\begin{figure}[ht!]
    \centering
    \begin{subfigure}[b]{0.75\linewidth}
        \centering
        \includegraphics[width=\linewidth]{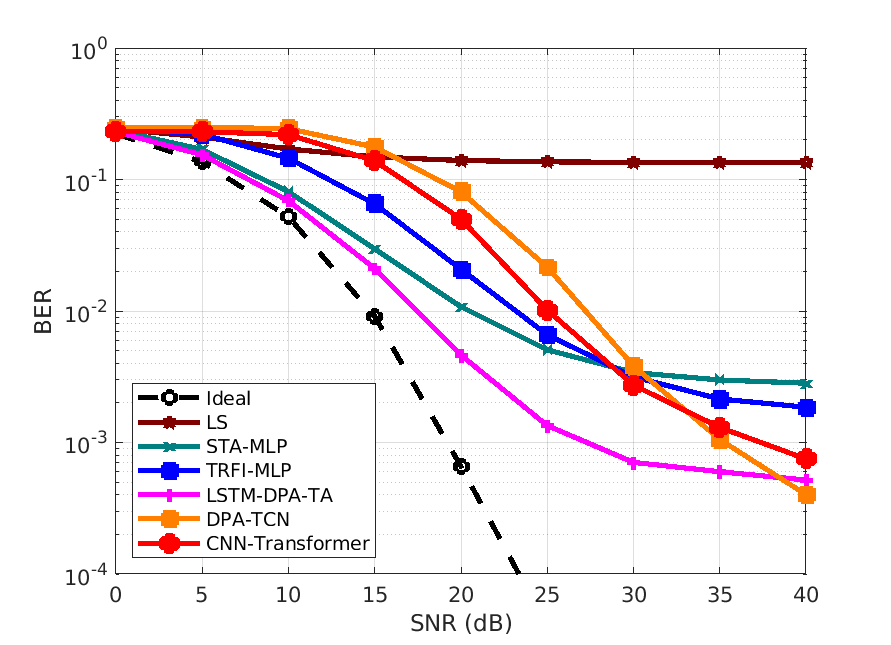}
        \caption{High SNR training dataset}
        \label{fig:high_snr_ber}
    \end{subfigure}
    \hfill
    \begin{subfigure}[b]{0.75\linewidth}
        \centering
        \includegraphics[width=\linewidth]{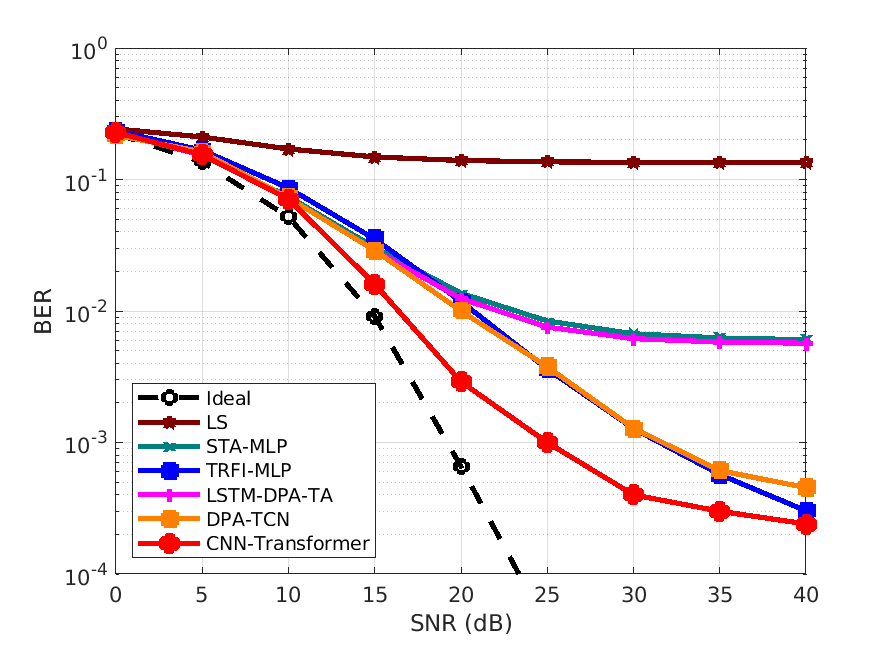}
        \caption{Mixed SNR training dataset}
        \label{fig:mixed_ber}
    \end{subfigure}
    \caption{Comparison of BER performance for various channel estimators trained on high SNR and mixed SNR datasets. \label{fig:ber}}
\end{figure}

%Figures \ref{fig:high_snr_ber}. and \ref{fig:mixed_ber}. 
Figure~\ref{fig:ber} presents the BER performance of the channel estimators trained using a high SNR dataset (40 dB) and a mixed SNR dataset, respectively. The ideal curve, represented by the dotted black line in these figures, represents the best possible performance in channel estimation, where the received signal is assumed to be perfectly equalised without any channel estimation errors. This curve serves as a theoretical lower bound for the BER, showing the performance in a scenario where only additive white Gaussian noise (AWGN) is present and the channel effects are perfectly compensated for. The closer the BER of a model is to the ideal curve, the better its performance in accurately estimating the channel and mitigating the effects of noise. LS serves as the lower performance baseline. 

In Figure \ref{fig:high_snr_ber}, we observe several key trends across models trained on an high SNR dataset: 
In the low SNR range (0-10 dB), the STA-MLP and the LSTM-DPA-TA models perform relatively well, maintaining a lower BER compared to other models. %The LSTM-DPA-TA model also exhibits strong performance, achieving the lowest BER across this range, demonstrating its suitability for low SNR conditions when trained on high SNR data.
As we move into the mid SNR range (15-25 dB), the CNN-Transformer and DPA-TCN models show an improvement, particularly from 15 dB onwards, with performance becoming more pronounced from 20 dB. %The improvement becomes more exponential from 20 dB. 
This improvement at high SNR highlights that these models have effectively adapted to high SNR conditions. %arnt to less noisy conditions but not high noisy conditions. 
%TRFI-MLP also exhibits a steady increase in BER with increasing SNR, although its performance lags behind that of CNN-Transformer and DPA-TCN. This also indicates the limitation of training NNs with the high SNR dataset as the TRFI-MLP struggles to adapt to noisy conditions.
The TRFI-MLP model also improves in this range, although it lags behind the CNN-Transformer and DPA-TCN models, indicating some limitations in its ability to adapt to low SNR conditions.
In the high SNR range (30-40 dB), the CNN-Transformer and DPA-TCN models continue to show further significant reductions in BER. The LSTM-DPA-TA model continues to outperform other models across all SNR levels, showing its effectiveness in less noisy conditions. In contrast, STA-MLP shows limited improvement beyond 20 dB, indicating that it struggles to capitalise on the reduced noise at higher SNR levels.

In the mixed SNR training scenario depicted in Figure \ref{fig:mixed_ber}, we observe that CNN-Transformer and DPA-TCN models exhibit excellent performance across the entire SNR range with CNN-Transformer outperforming all models.  %This suggests that these models are well-suited for a wide range of channel conditions when trained on mixed SNR datasets.
The LSTM-DPA-TA and STA-MLP models perform well at lower SNR levels (0 to 15 dB) but show reduced performance as the SNR increases, indicating diminished capabilities in high SNR conditions when trained on mixed SNR data. We can also observe the improved performance of the TRFI-MLP estimator across the entire SNR range tested compared to its performance when trained on the high SNR dataset. The unsatisfactory performance by LSTM-DPA-TA and STA-MLP indicates that while these models can handle varying noise conditions, they may not leverage mixed SNR training as effectively as TRFI-MLP, CNN-Transformer and DPA-TCN.

Models trained on mixed SNR data exhibit lower BER even in low-SNR scenarios, compared to those trained on high SNR data. Mixed SNR training appears to be beneficial for models such as TRFI-MLP, TCN-DPA, and CNN-Transformer, which show a significant improvement in low SNR conditions than when trained on high SNR. In contrast, the LSTM-DPA-TA and STA-MLP models demonstrate better performance when trained on a high SNR dataset, suggesting that these models make use of the better channel statistics available at high SNR. This is analysed in more detail below.

\subsection{Difference in BER Between Models Trained on Mixed SNR and High SNR Datasets}
\begin{figure}[ht!]
    \centering
    \includegraphics[width=0.8\linewidth]{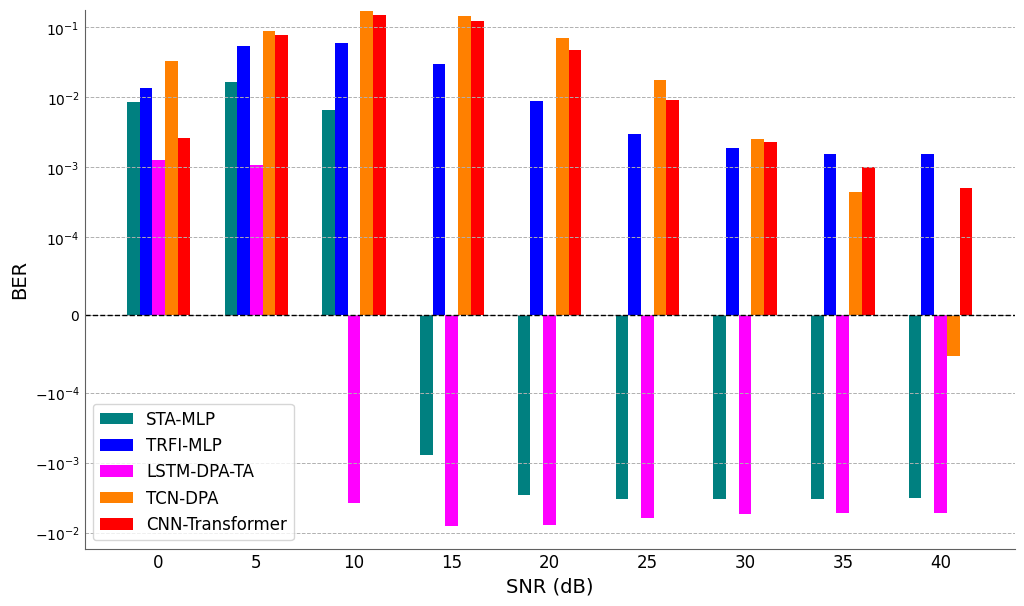}
    \caption{Difference in BER between models trained on mixed SNR and high SNR datasets.}
    \label{fig:delta_ber}
\end{figure}

Figure~\ref{fig:delta_ber} illustrates the difference in BER between models trained on high SNR datasets and those trained on a mixed SNR dataset. The graph clearly shows how the training dataset influences the performance of various models across different SNR ranges. In the low SNR range (0-10 dB), the CNN-Transformer and TCN-DPA models exhibit the highest positive delta, peaking around 10 dB. This indicates a substantial improvement in performance when these models are trained on a mixed SNR dataset compared to a high SNR dataset. A positive delta suggests that mixed SNR training better equips these models to generalise in low SNR conditions. TRFI-MLP also shows a noticeable positive delta, although less pronounced, indicating that mixed SNR training offers some advantages in low SNR environments. In contrast, the LSTM-DPA-TA and STA-MLP models display a slightly negative delta across this range, implying that training on high SNR datasets offers a marginal performance advantage in these specific models. 

As the SNR increases (15-25 dB), the positive delta values for the CNN-Transformer and TCN-DPA models start to decrease but remain positive, particularly around 15 dB. This indicates that while the benefits of mixed SNR training diminish slightly, these models still gain performance advantages in this SNR range. TRFI-MLP continues to maintain positive delta values, demonstrating that mixed SNR training consistently improves its performance at varying SNR levels. Meanwhile, the LSTM-DPA-TA model continues to show a negative delta, particularly at the higher end of the mid-SNR range, around 20-25 dB. Similarly, the STA-MLP model exhibits a slight negative delta in this range.

In the high SNR range (30-40 dB), the delta values for most models, including the CNN-Transformer, TCN-DPA, and TRFI-MLP, approach zero. This suggests that as the SNR increases and the impact of noise decreases, the performance difference between models trained on high SNR and mixed SNR datasets becomes negligible. %The LS, STA-MLP, and LSTM-DPA-TA models exhibit very low or slightly negative delta values, indicating that their performance is largely unaffected by the training dataset's SNR range in higher SNR environments.

\section{Conclusion} 
This study investigated the effectiveness of training NN-based channel estimators using mixed SNR datasets compared to high SNR datasets. Our results demonstrate that training with mixed SNR data significantly improves the generalisation of various estimators, especially in low SNR conditions. In particular, models such as the CNN-Transformer, DPA-TCN and TRFI-MLP exhibited substantial improvements in BER across the entire SNR range when trained on mixed SNR data compared to when trained on high SNR. Among the models tested, the CNN-Transformer, when trained on mixed SNR data, outperformed other estimators, including the current state-of-the-art LSTM-DPA-TA. However, it is important to note that some models, such as LSTM-DPA-TA and STA-MLP, showed reduced performance when trained on mixed SNR data. 
Mixed SNR training improves performance and generalisation across SNR levels for some models. The exact reason why only certain models benefit remains unclear and requires further investigation. 

These results indicate the importance of considering the SNR range as an important hyperparameter during training, rather than following the current practice of using only high SNR training data. The channel model used in this study is an example where the impact of mobility on the channel is more significant compared to other channels with lower mobility. As a result, the tested models should be able to generalise well to those channels. This will be verified in future research that will investigate the impact of mixed SNR training on other vehicular channel models.

\subsubsection{Acknowledgements}
The authors are grateful to NITheCS and the Telkom CoE at NWU for their support of this research.

 \bibliographystyle{splncs04}
 \clearpage
 \bibliography{references}

\end{document}